\documentclass[a4paper]{article}

\usepackage{INTERSPEECH2019}
\usepackage{bm}
\usepackage{amsmath}
\usepackage{amsfonts}
\usepackage{amssymb}
\usepackage{calrsfs}
\usepackage{algorithm}
\usepackage{algorithmic}
\usepackage{graphicx,subfigure}

\title{One-vs-All Models for Asynchronous Training: An Empirical Analysis} 
\name{Rahul Gupta, Aman Alok, Shankar Ananthakrishnan}

\address{Amazon.com, Cambridge, MA-USA}
\email{gupra@amazon.com, alokaman@amazon.com, sanantha@amazon.com}

\begin{document}

\maketitle
\begin{abstract}
Any given classification problem can be modeled using multi-class or One-vs-All (OVA) architecture. An OVA system consists of as many OVA models as the number of classes, providing the advantage of asynchrony, where each OVA model can be re-trained independent of other models. This is particularly advantageous in settings where scalable model training is a consideration (for instance in an industrial environment where multiple and frequent updates need to be made to the classification system). In this paper, we conduct empirical analysis on realizing independent updates to OVA models and its impact on the accuracy of the overall OVA system. Given that asynchronous updates lead to differences in training datasets for OVA models, we first define a metric to quantify the differences in datasets. Thereafter, using Natural Language Understanding as a task of interest, we estimate the impact of three factors: (i) number of classes, (ii) number of data points and, (iii) divergences in training datasets across OVA models; on the OVA system accuracy. Finally, we observe the accuracy impact of increased asynchrony in a Spoken Language Understanding system. We analyze the results and establish that the proposed metric correlates strongly with the model performances in both the experimental settings. 
\end{abstract}

\section{Introduction}
Any classification problem can be solved in either a multi-class setup or an One-vs-All (OVA) setup \cite{bishop2006pattern}.
While similar modeling architectures can be used for both the setups, certain inherent differences exist between them. 
Firstly, a multi-class modeling with $K$ classes yields a single model, while an OVA modeling yields $K$ different models.
Given a test instance, the multi-class model outputs a probability vector with entries for each class. 
On the other hand, each individual OVA model yields a scalar probability for membership of the test instance to the corresponding class. 
Typically, class probabilities from the multi-class system sum to 1, while that may not be true for an OVA system.
Another important distinction between multi-class and OVA systems is that constituent OVA models in the system can be updated asynchronously.
Alterations to the model architecture, as well as to the training data for each OVA model can be realized asynchronously without an exhaustive overhaul of the classification system - an option not available in multi-class system. 
This property of OVA systems is particularly attractive when scalability and flexibility of models is a constraint. 
For instance, if two model updates are requested, they can potentially be recognized asynchronously with an OVA system (e.g. when they pertain to different OVA models in the system). 
Alternately, in a multi-class system, a full-scale retraining is required either by queuing the updates or merging them.
The goal of this paper is to explore this particular property of OVA systems.  
We are interested in the impact of factors such as dataset divergence across OVA models, dataset size and number of classes on the overall OVA system accuracy. 

Rifkin and Klautau \cite{rifkin2004defense} present an in-depth theoretical and empirical analysis on the performance of OVA models.
They argue that OVA systems can perform as well as any other multi-class classification or an all-vs-all classification setup. 
Following this, other works have presented further modifications to the OVA modeling scheme \cite{beygelzimer2005weighted, galar2011overview, dinh2015learning} for an improved performance. 
Further empirical evidence showcasing the utility of OVA models is shown in applications such as image classification \cite{joutsijoki2011comparing, mathur2008multiclass}, defect detection \cite{ng2014one} and data stream classification \cite{hashemi2009adapted}. 
However, most of this work is focused on performance of OVA models against other modeling approaches.
Our work, on the other hand, focuses on the asynchrony aspect of OVA models, which allows training and maintaining each OVA model independent of others.
We conduct experiments on Natural Language Understanding (NLU) tasks, specifically a domain classification task and an SLU task to obtain domain-intent-entity combinations \cite{sarikaya2017technology}. 
Although both multi-class \cite{robichaud2014hypotheses, sarikaya2016overview} and OVA designs \cite{su2018re} exist for such tasks, a comparison between the two systems with asynchrony capabilities added in the OVA setup has not been explored. 

We conduct two sets of experiments in this paper. 
First, a synthetic setup evaluates the impact of asynchronous training of OVA models on the overall performance of the OVA system.
Using domain classification as the NLU task of interest, we vary the number of data-points, classes and asynchrony in training datasets in an OVA system and compare the model performance against a multi-class setup. 
We then conduct a second experiment on a large scale SLU system and evaluate the impact of asynchrony on the system.
While the first experiment is a simulation, in the second experiment, dataset asynchrony in training OVA models is more naturalistic. 
This is due to the fact that datasets in SLU systems evolve continually due to availability of more data, modification of existing data and even deprecation. 
Another objective of ours is to propose a measure that can quantify the asynchrony in the OVA system and can be used to assess the health of the system. 
We propose an asynchrony metric $\alpha$ in the Section~\ref{sec:metric}. 
In case of synthetic experiments, the proposed metric obtains a correlation above 0.87 with the absolute differences in error rates of multi-class and the OVA system.
For the real world SLU system, the correlation between relative degradation (between an OVA setup with asynchrony and a multi-class setup) in an Semantic Error Rate metric and metric $\alpha$ equals 0.84. 
We present further analysis on the performance trends between OVA and multi-class systems and in the next section, we introduce the notations used in this paper.
 
\section{Setup of an OVA system}
\label{sec:notation}
Consider a multi-class classification task with $K$ classes $\{y_1, y_2, .., y_K\}$, where data-points belonging to the $k^\text{th}$ classes are part of the dataset $\mathbb {\bm D}_k$.
An OVA classification setup for this task will consist of $K$ models, where the model ($\bm M_k$) is trained to predict confidence scores for the class $k$ against all other classes.
The model is trained with data-points from the set $\mathbb {\bm D}_k$ as positive samples, while data-points from the other sets $\mathbb {\bm D}_l; l \neq k$ are pooled to create negative samples. 
In our setting, we assume that while $\bm M_k$ is retrained every-time the positive samples $\mathbb {\bm D}_k$ are updated, re-training may not be performed when other datasets $\mathbb {\bm D}_l; l \neq k$ are updated/modified.
This allows for an asynchronous update of each of the models $\bm M_1, .., \bm M_K$. 
We denote the negative samples from $\mathbb {\bm D}_l$ used to train $\bm M_k$ as $\mathbb {\bm D}_l^k$. 
Hence a given model $\bm M_k$ is trained on a collection of datasets $\{\mathbb {\bm D}_1^k, .., \mathbb {\bm D}_k, .., \mathbb {\bm D}_{K}^k\}$, where only $\mathbb {\bm D}_k$ contributes positive samples.

\subsection{Defining an asynchrony measure across OVA systems}
\label{sec:metric}
Expectedly, re-training each model $\bm M_k, k=1, .., K$ every-time any subset of the datasets $\mathbb {\bm D}_k, k=1, .., K$ is updated would yield the best results. 
However, since we assume that each of these models is maintained and updated asynchronously, at any given point in time the constituent OVA models may individually be trained on slightly different datasets. 
In order to quantify the differences in datasets used to train each OVA model $\bm M_k$, we define an asynchrony measure $\alpha$.
Assuming that the OVA system will take a performance hit with the introduction of asynchrony in training data for each OVA model, the metric $\alpha$ is intended to correlate well with the performance degradation and provide an overall estimate of the health of the OVA model. 
Algorithm 1 provides the pseudo-code to compute this metric. 

\begin{algorithm}[t]
\begin{algorithmic}
\FOR{k=1, .., K}
  \STATE {\bf Obtain} all copies of a given dataset $\mathbb {\bm D}_k: \mathbb {\bm D}_k^l, l \in \{1,..,K\} - k $ 
  \STATE {\bf Obtain} sentence embeddings for all utterances in $\mathbb {\bm D}_k$ 
  \STATE {\bf Model} the distribution of sentence embeddings from $\mathbb {\bm D}_k$ using KDE. Resulting PDF is denoted as $P_k$. 
  \FOR{each copy $\mathbb {\bm D}_k^l$ of the set $\mathbb {\bm D}_k \in \{\mathbb {\bm D}_k^1,.., \mathbb {\bm D}_k^K$\}}
    \STATE {\bf Obtain} sentence embeddings for utterances in $\mathbb {\bm D}_k^l$. 
    \STATE {\bf Model} the distribution of sentence embeddings from $\mathbb {\bm D}_k^l$ using KDE. PDF obtained from $\mathbb {\bm D}_k^l$ is denoted as $P_k^l$.
    \STATE {\bf Obtain} the log-likelihood ratio: 
    \STATE $a_k^l = \frac{1}{|\mathbb {\bm D}_k|}\sum_{x_i \in \mathbb {\bm D}_k} \log \frac {P_k^l (\bm x_i)} {P_k(\bm x_i)}$ 
  \ENDFOR
  \STATE $a_k =$ Mean$(a_k^l), l \in \{1,..,K\} - {k}$ 
\ENDFOR
 \STATE Asynchrony metric $\alpha = \sum_{k=1}^K a_k$
 \end{algorithmic}
\caption{Algorithm to compute the asynchrony metric}
\end{algorithm}

Note that for any given dataset $\mathbb {\bm D}_k$, $K$ copies $\mathbb {\bm D}_k^1,.., \mathbb {\bm D}_k, \mathbb {\bm D}_k^K$ exist, ($\mathbb {\bm D}_k^1, \mathbb {\bm D}_k^2, ..$ are used  to train  $\bm M_1, \bm M_2, ..$ as negative samples, while $\mathbb {\bm D}_k$ is used to train $\bm M_k$ as positive samples). 
The algorithm first obtains a Probability Distribution Function (PDF: $P_k$) for sentence embeddings obtained from the dataset $\mathbb {\bm D}_k$. 
We average word embeddings to obtain sentence embeddings. Word embeddings are trained on an Alexa corpus with about 10M sentences using the skip-gram objective \cite{mikolov2013distributed}.
The PDF $P_k$ is estimated using Kernel Density Estimation (KDE) \cite{bishop2006pattern} on the sentence embeddings.
The hyper-parameters for KDE are heuristically chosen: we obtain KDE using a Gaussian kernel with a diagonal variance of .01.
Similarly, we obtain a PDF $P_k^l$ on a copy of the dataset $\mathbb {\bm D}_k^l$ by performing KDE on the sentence embeddings.
Finally, we obtain a log-likelihood ratio $a_k^l$ of the sentence embeddings in the dataset $\mathbb {\bm D}_k$ using $P_k$ and $P_k^l$ (The log likelihood is normalized with the size of the dataset $|\mathbb {\bm D}_k|$). 
We average $a_k^l$ over all the copies of the dataset (barring $\mathbb {\bm D}_k$ itself) to obtain $a_k$. 
Finally, the metric $\alpha$ is obtained as sum of the averages $a_k$ computed for each dataset $\mathbb {\bm D}_k$. 

The metric $\alpha$ captures the dissimilarity between a dataset and it's copies by comparing the distribution of constituent utterances in the latent sentence embedding space.  
Copy $\mathbb {\bm D}_k^l$ can be different from $\mathbb {\bm D}_k$ in two ways.
Either, the copy can be differently distributed (due to a biased sampling or major modification of $\mathbb {\bm D}_k$ post the creation of copy) or, it can be similarly distributed but sub-sampled.
In either case, KDE will yield a different PDF for the copy $\mathbb {\bm D}_k^l$, reflecting the asynchrony in the metric $\alpha$.
Note that we also sum the values $a_k$ per dataset, hence an increase in metric $\alpha$ is expected as $K$ increases.
This is done intentionally to capture the intuition that OVA system performance will degrade as the number of OVA models in the system increases.
In the next section, we use the asynchrony metric $\alpha$ and correlate it with the performances of OVA system as a function of dataset size, number of classes $K$ and asynchrony amongst the datasets.

\section{Analysis on the behavior of one-vs-all systems}
\label{sec:synthetic}
In this experiment, we compare the performance gap between an OVA system and a multi-class system, as more asynchrony is injected in the OVA system.  
We conduct an analysis on the behavior of OVA systems as a function of three factors: (i) the asynchrony between a dataset and it's copies, (ii) the size of the datasets and, (iii) number of one-vs-all systems (equivalently number of classes $K$).
We perform simulations perturbing these factors in a domain classification task catered towards a Spoken Language Understanding (SLU) system. 
We first describe the dataset used for this analysis followed by the model description and a description of the simulation experiments. 

\subsection{Dataset}
\label{sec:dataset}
We use a set of $\sim$1.5M utterances derived from user requests directed towards Alexa devices. 
Each utterance is annotated to belong to an Alexa enabled domain (out of a set of 23 domains). 
These utterances are divided into training, development and testing sets in a speaker independent manner (that is, a given speaker only has all his/her utterances in either training, development or testing set) using 8:1:1 ratio.
An OVA system trained on this data will have $K=23$ models.

\begin{figure*}[t]
\centering
\subfigure[Asynchrony factor]{
\includegraphics[width=.32\textwidth]{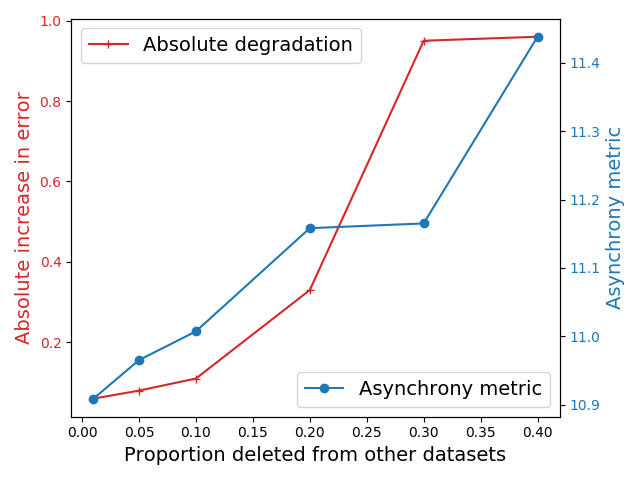}
\label{fig:async}}
\subfigure[Number of data-points]{
\includegraphics[width=.32\textwidth]{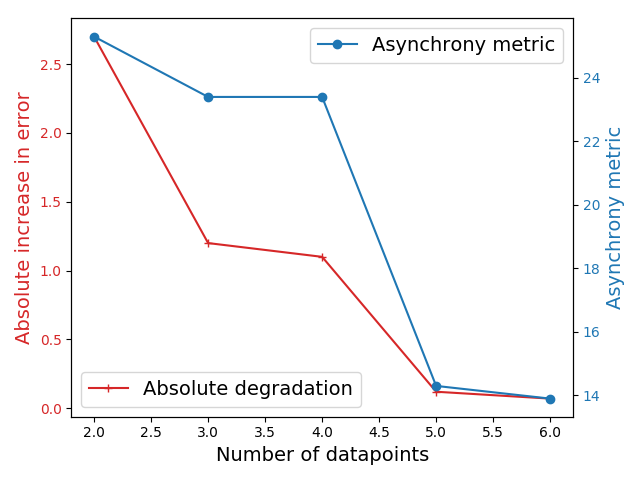}
\label{fig:datapoints}}
\subfigure[Number of classes]{
\includegraphics[width=.32\textwidth]{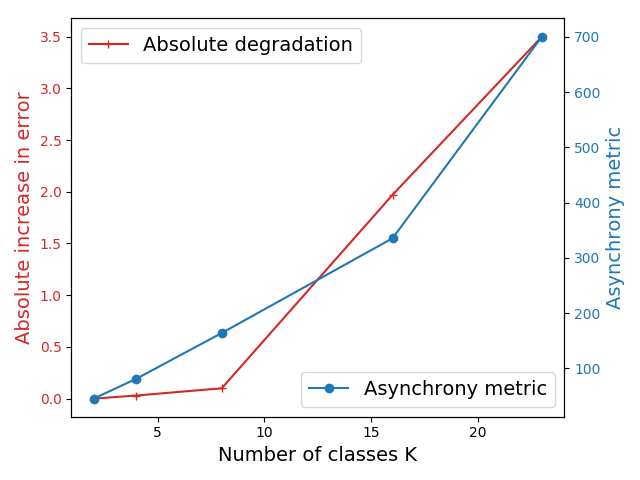}
\label{fig:classes}}
\vspace{-3mm}
\caption{Figures depicting absolute increase in OVA system classification error rate as compared to a multi-class system: (a) performances gap as an OVA model $\bm M_k$ is trained with fewer negative samples borrowed from $\mathbb {\bm D}_l, l \neq k$, (b) performance gap with a fixed sub-sampling from other datasets $\mathbb {\bm D}_l, l \neq k$ but with increasing overall dataset size and, (c) performance gap as number of classes increase with a fixed sub-sampling and overall dataset size.}
\label{fig:whatever}
\end{figure*}

\subsection{Model architecture}
For our simulations, we use a model architecture with the following components: an embedding layer, an encoding layer and a fully connected layer.
The output of the fully connected layer is a single class probability in case of OVA models, while it is a $K$ dimensional vector in case of multi-class classification. 
The embedding layer maps each word in input utterance to a 256 dimensional vector (the word embeddings are pre-trained using an Alexa corpus consisting of $\sim$ 10M utterances).
The word embeddings are then passed through an encoding layer.
We use a Convolutional Neural Network (CNN) encoder with 1-dimensional filters that act on a window of word embeddings at a given time \cite{hu2014convolutional}.
The outputs of the CNN layer are then passed through a fully connected layer for final classification.
We tune the model hyper-parameters on a development set for each experiment (in case the training set is down-sampled for an experiment, the development set is also down-sampled. If the size of training set is set to less than 10k, size of development set is kept same as the size of the training set). 

\subsection{Simulation experiment setup}
We describe our experimental setup to assess the impact of asynchrony across datasets, number of data-points and the number of classes on an OVA system performance below.

\subsubsection{Asynchrony factor}
In order to understand the impact of data asynchrony in OVA systems, we train OVA systems with an increasing amount of data asynchrony between a dataset $\bm D_k$ and it's copies $\bm D_k^l$ (number of classes $K$ and overall size of the dataset is not altered in this experiment).
In order to obtain negative samples, we randomly sub-sample a fraction of utterances from the datasets $\mathbb {\bm D}_{l}; l \neq k$ to train $\bm M_k$.
We decrease the sampled fraction and compare the performance of the OVA system against a multi-class system trained on the collection of datasets $\mathbb {\bm D}_k, k=1,..,K$ (note that the multi-class classification system is trained on original copies of datasets without any sub-sampling).
%The class assignment for the OVA system is made based on the arg-max operation over the class specific confidences, as returned by each individual model $\bm M_k$.
Figure~\ref{fig:async} presents the absolute difference in performances between the multi-class and OVA systems. 
We also compute the value of the asynchrony metric $\alpha$ for each sampling from the other domain's datasets and observe the metric's correlation with the performance difference between multi-class and OVA systems. 

\subsubsection{Number of data points}  
In this experiment, we evaluate the impact of dataset size on the performance of the OVA systems.
Keeping the fraction of utterances sub-sampled from $\mathbb {\bm D}_l; l \neq k$ to train $\bm M_k$ constant at 30\%, we increase the size of the datasets $\mathbb {\bm D}_1, .., \mathbb {\bm D}_K$ themselves.  
The overall dataset described in Section~\ref{sec:dataset} is sampled to create copies of datasets $\mathbb {\bm D}_1, .., \mathbb {\bm D}_K$ with sizes increased from 100 to 1M in a logarithmic scale.
The architecture of the OVA models and the multi-class classifier is kept same for each comparison. 
However as the datasets size increases, the models are fine-tuned to higher complexity models (e.g. with more convolutional filters and hidden layers). 
We also compute the asynchrony measure $\alpha$ for each size and Figure~\ref{fig:datapoints} presents the absolute difference in performances between the multi-class and OVA systems. 

\subsubsection{Number of classes}
Finally, in order to estimate the impact of the number of classes on performance of OVA systems, we perform a simulation with increasing number of classes $K$.
We randomly merge the constituent 23 classes in the domain classification dataset to conduct experiments with number of classes $K$ set to 2, 4, 8, 16 and 23. 
We again randomly sub-sample a constant fraction of utterances from $\mathbb {\bm D}_{k^\prime}; k \neq k^\prime$ to train $\bm M_k$ for OVA system, while the multi-class model is trained on the combined dataset.
Figure~\ref{fig:classes} plots the asynchrony measure along with the absolute performance difference of the OVA systems and a multi-class system. 

\subsection{Discussion}
For the experiment with increasing asynchrony factor, we observe that the performance gap between the OVA system and multi-class system increases more rapidly as we delete more than 10\% of the negative samples from other datasets.
Similarly, as we increased the number of classes beyond 8, the difference in performance accelerates as more and more classes are introduced. 
This suggests that the performance gap increases non-linearly as more asynchrony or more number of classes are introduced in an OVA system.
For the experiment with increasing dataset size, the regression appears to increase non-linearly with log size of the dataset.
Hence the regression severity due to asynchronous training can be reduced by increasing the data sizes.
This is intuitive as random sub-sampling should not alter the distribution of larger datasets. 

We also observe that the metric $\alpha$ correlates well with the performance of the OVA system.
With the available (handful) experimental data-points with simulated increase in asynchrony factor, dataset size and number of classes, we obtain correlations of 0.86, 0.88 and 0.97 (respectively) between the metric $\alpha$ and performance gaps.
The correlation with the number of classes is particularly high as just the correlation between number of classes and the performance gap is 0.96 and the metric $\alpha$ marginally outperforms this correlation. 
This observation suggests that this metric could be used as a surrogate to estimate the health of the OVA system. 
%If the metric $\alpha$ trips a particular threshold, it can be used as an indicator that datasets across the OVA models need to be synchronized and retrained.
In the next section, we perform experiments on an SLU system setup in OVA fashion and conduct similar analysis.

\section{Analysis on an One-vs-all SLU system}
In this section, we conduct an analysis on a more complex real world OVA system: an SLU system consisting of multiple components operating together.
We use the SLU model setup developed by Su et al. \cite{su2018re}, where each domain in the NLU system contains a set of four models: (i) a Domain Classifier (DC), (ii) an Intent Classifier (IC), (iii) a Named Entity Recognizer (NER) and, (iv) a re-ranker.
The DC and re-ranker components are setup in OVA fashion and they produce domain specific confidences for domain and NLU hypotheses.
IC and NER models are trained only on domain specific data (no negative data is borrowed from other domains).
The setup of these models is described in more detail in \cite{su2018re}, and the set of DC, IC, NER and re-ranker models in each domain is referred to as $\bm M_k$ in this section. 

In the NLU model setup, our experimental setting assuming independent model training fits naturally. 
It is desirable to have this flexibility to allow independent maintenance of domain specific models $\bm M_k$ and the corresponding positive data $\mathbb {\bm D}_k$.
The additive value of these experiments against the synthetic experiments in Section~\ref{sec:synthetic} is that the source of discrepancies in determining datasets for training each set of models $\bm M_k$ are realistic. 
The datasets intrinsic to each domain $\mathbb {\bm D}_1, .., \mathbb {\bm D}_K$ manifest independently due to availability of new data, data deprecation and/or modification.
Our experimentation also covers instances where data from a given domain may not even be available to sample negative data for another domain's models (for instance, if the other domain's models have not been re-trained since the addition of a new domain). 
We describe the setup of our analysis in the next section. 

\subsection{Experimental setup on the SLU system}
\begin{figure}[t]
\centering
\includegraphics[width=.4\textwidth]{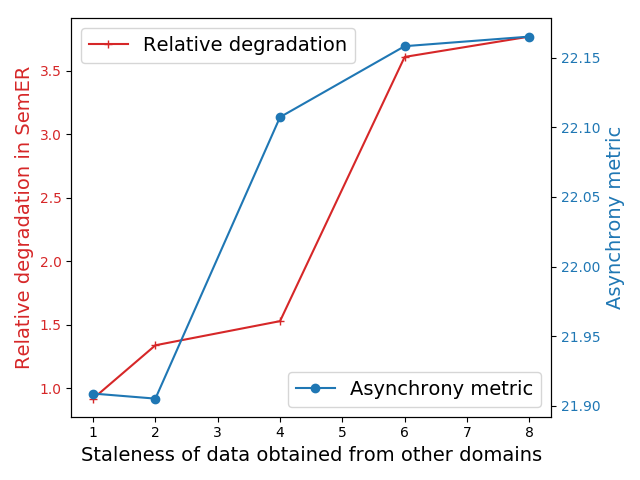}
\vspace{-4mm}
\caption{Relative SemER degradation in an OVA SLU system compared to a multi-class SLU system. Staleness is determined in terms of month time units.}
\vspace{-5mm}
\label{fig:slu}
\end{figure}

The SLU system in our experiments consists of 23 domains and we use the same data source as described in \cite{su2018re}. 
The models $\bm M_k$ for each domain is trained on a collection of datasets $\{ \mathbb {\bm D}_1^k, .., \mathbb {\bm D}_k, .., \mathbb {\bm D}_K \}$, where $\mathbb {\bm D}_l^k$ could be an outdated copy of data $\mathbb {\bm D}_l$ from the domain $l$.
We introduce asynchrony in the SLU model training, where the negative data $\mathbb {\bm D}_l^k$ borrowed from other domains could be upto $N$ months old.
As the value of $N$ increases, older copies from other domains get incorporated in training the models $\bm M_k$.
We track the performance of the SLU system as $N$ increases using the Semantic Error Rate (SemER) metric defined in \cite{su2018re}. 
The metric quantifies the performance of the SLU system by checking for accuracy of the intents and slots produced by the SLU system. 
The performance of the OVA SLU system is compared against a multi-class SLU system with multi-class DC and reranker trained on a synchronized collection of datasets $\{ \mathbb {\bm D}_1^k, .., \mathbb {\bm D}_k, .., \mathbb {\bm D}_K \}$ (we report relative degradations in this experiment due to lower error rates). 
Figure~\ref{fig:slu} presents the performance degradations between the two systems. 
We also plot the asynchrony metric $\alpha$ along with the SemER value to observe their correlation. 

\subsection{Discussion}
From the results, we observe that the value of relative degradation in a SLU system also increases rapidly beyond $N=4$. 
This is consistent with the observations in synthetic experiments with increased asynchrony amongst datasets. 
We also note that as $N$ increased in value, there were instances when a particular model $\bm M_k$ was trained without any negative instance from a set of domain newly launched in the past $N$ months.

The correlation of the metric $\alpha$ with the SemER value degradation is also high (0.84), suggesting the applicability of this metric in real world systems.
In a real world setting, we suggest an operational model where a baseline value for metric $\alpha$ can be computed for a given acceptable state of the OVA models.
Metric $\alpha$ can be recomputed each time any model in the OVA system is updated and if the relative degradation in $\alpha$ exceeds a certain threshold, that may warrant a re-syncing of datasets across models and retraining them. 
We note that the metric $\alpha$ is designed to provide the right assessment of the health of an OVA SLU system as the dataset sized vary and more domains are added.

\section{Conclusion}
OVA systems have been compared to a multi-class system for accuracy in several previous works \cite{rifkin2004defense}.
In this work, we explore one specific property of OVA systems, where each OVA model can be updated asynchronous of other OVA models.
Our application of choice in this paper are SLU tasks and we propose a metric that can quantify asynchrony amongst SLU datasets by modeling their distribution in a latent space (sentence embeddings). 
We conduct synthetic experiments perturbing the size of the datasets, number of classes as well as the asynchrony in constructing dataset for each OVA model training. 
We observe the trends in performance degradation as compared to a multi-class model (trained on the synchronized data) and obtain good correlations of the asynchrony metric $\alpha$ with the performance degradation.
Finally, we conduct experiments on a real world SLU system with a more natural setting where dataset asynchrony is introduced based on the state of datasets at a given point in time.
We make similar observations as the synthetic setting and obtain a good correlation between the performance degradation and the metric $\alpha$.

In the future, we aim to further fine tune the metric $\alpha$ for better correlation with the model's performance \cite{lindsey1974comparison, cha2007comprehensive}.  
For instance, we aim to experiment with adding a transformation to the sentence embeddings before PDF estimation to enhance differences in dataset that can lead to a higher performance degradation.
We also aim to further experiment with mechanisms that can be used to gate asynchrony in OVA systems using the metric $\alpha$ and provide signals to inform model retraining.
This also involves deriving a theoretical relationship between the model performances and the metric $\alpha$ (beyond the empirical experiments in our work).
Finally, we also aim to extend the analysis to tasks beyond Natural Language Understanding.

\bibliographystyle{IEEEtran}

\bibliography{mybib}

\end{document}